\documentclass[final]{cvpr}

\usepackage{times}
\usepackage{epsfig}
\usepackage{graphicx}
\usepackage{amsmath}
\usepackage{amssymb}
\usepackage{xcolor}
\usepackage{mynotation}
\usepackage{booktabs}
\usepackage[caption=false]{subfig}

\usepackage[pagebackref=true,breaklinks=true,colorlinks,bookmarks=false]{hyperref}

\newcommand{\parsection}[1]{\vspace{0.5mm}\noindent\textbf{#1:}~}

\usepackage{capt-of,etoolbox}
\makeatletter
\apptocmd\@maketitle{{\introfig{}\par}}{}{}
\makeatother

\newcommand{\weightpred}{W}
\newcommand{\flowestimator}{F}
\newcommand{\warp}{\phi}
\newcommand{\fusionweightraw}{\tilde{w}}
\newcommand{\fusionweight}{w}
\newcommand{\bimage}{b}
\newcommand{\bimagepacked}{\tilde{\bimage}}
\newcommand{\encfeat}{e}
\newcommand{\warpedfeat}{\tilde{\encfeat}}
\newcommand{\warpedfeatproj}{\warpedfeat^p}
\newcommand{\mergedfeat}{\hat{\encfeat}}
\newcommand{\flow}{f}
\newcommand{\flowfeatures}{\hat{\flow}}
\newcommand{\gtimage}{y_\text{GT}}
\newcommand{\ccm}{C}

\begin{document}

\title{Deep Burst Super-Resolution}

\author{Goutam Bhat \qquad Martin Danelljan \qquad Luc Van Gool \qquad Radu Timofte\\
	Computer Vision Lab, ETH Zurich, Switzerland\\
}

\newcommand{\introfig}{
\centering%
	\newcommand{\wid}{0.96\textwidth}%
	\includegraphics*[trim = 0 10 0 0, width = \wid]{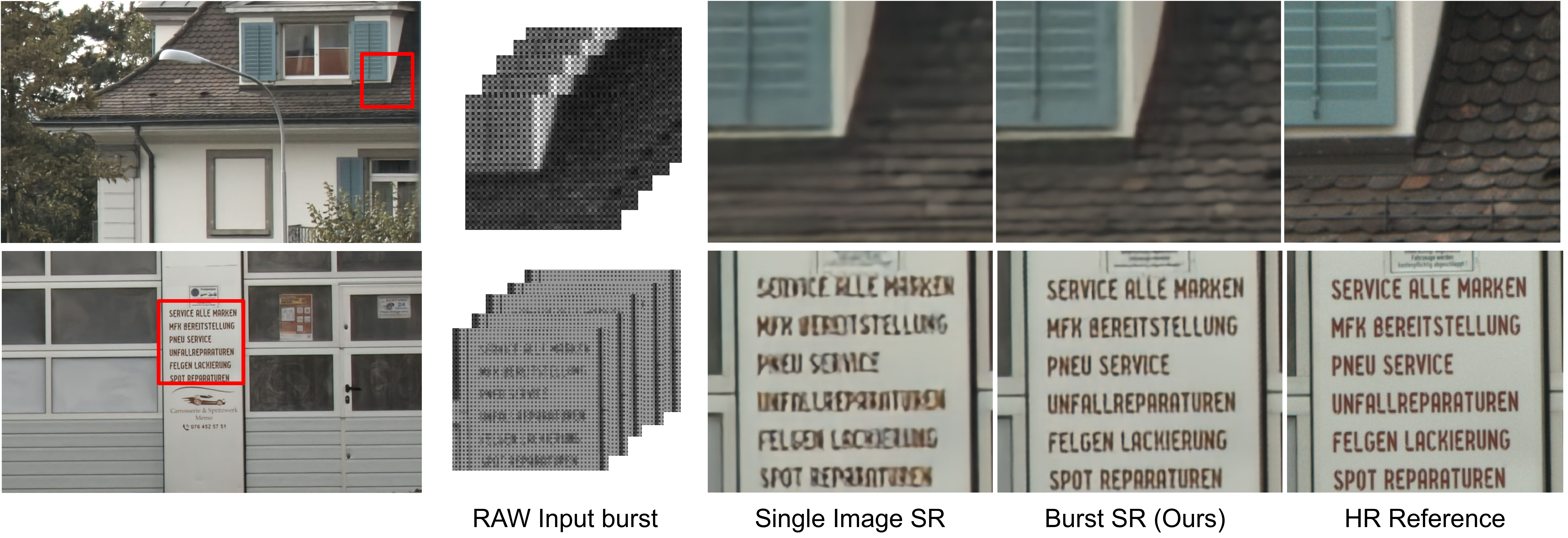}\hspace{1mm}
	\vspace{0mm}
	\captionof{figure}{Our network generates a super-resolved RGB image from an input burst consisting of multiple noisy RAW frames.
	In contrast to the single image baseline, our approach combines information from multiple frames to obtain a more detailed reconstruction of the scene. The results shown are for super-resolution by a factor of 4.
	}\vspace{0mm}%
	\label{fig:intro}
\vspace{8mm}}

\maketitle

\begin{abstract}

While single-image super-resolution (SISR) has attracted substantial interest in recent years, the proposed approaches are limited to learning image priors in order to add high frequency details.
In contrast, multi-frame super-resolution (MFSR) offers the possibility of reconstructing rich details by combining signal information from multiple shifted images. This key advantage, along with the increasing popularity of burst photography, have made MFSR an important problem for real-world applications.

We propose a novel architecture for the burst super-resolution task. Our network takes multiple noisy RAW images as input, and generates a denoised, super-resolved RGB image as output. This is achieved by explicitly aligning deep embeddings of the input frames using pixel-wise optical flow. The information from all frames are then adaptively merged using an attention-based fusion module. In order to enable training and evaluation on real-world data, we additionally introduce the \emph{BurstSR} dataset, consisting of smartphone bursts and high-resolution DSLR ground-truth. We perform comprehensive experimental analysis, demonstrating the effectiveness of the proposed architecture.
\end{abstract}

\section{Introduction}
Super-resolution (SR) is the task of generating a high-resolution (HR) image, given one or several low-resolution (LR) observations. It is a widely studied problem~\cite{Deudon2020HighResnetRF,Dong2016ImageSU,Johnson2016PerceptualLF,Kim2016AccurateIS,Lai2017DeepLP,Ledig2017PhotoRealisticSI,Lim2017EnhancedDR,srflow,Tai2017ImageSV,Wang2018ESRGANES,Wronski2019HandheldMS,Yu2016UltraResolvingFI,Zhang2018ResidualDN} with numerous practical applications. In recent years, the SR community has mainly focused on the single image super-resolution (SISR) task, where an HR image is estimated from a single LR input. Due to the ill-posed nature of the SISR problem, these methods are limited to adding high frequency details through learned image priors.

The multi-frame super-resolution (MFSR), on the other hand, aims to reconstruct the original HR image using multiple LR images. If the input images have sub-pixel shifts with respect to each other, due to \eg camera motion, they provide different LR samplings of the underlying scene. MFSR approaches can thus exploit this additional signal information to generate a higher quality image, compared to the SISR approaches (see Fig.~\ref{fig:intro}). The MFSR problem naturally arises in the increasingly popular mobile burst photography, where the images have different sub-pixel shifts due to natural hand tremors~\cite{Wronski2019HandheldMS}. This opens up the possibility of using MFSR to overcome the resolution constraint in mobile cameras imposed by the cost and size restrictions.

Despite the aforementioned advantages, MFSR has received little attention in recent years.
This is in stark contrast to SISR, where deep learning has led to significant advancements in SR performance.
Compared to the SISR case, the MFSR problem imposes significant challenges when developing deep learning based solutions. Firstly, a MFSR architecture must be able to align the noisy input frames with sub-pixel accuracy in order to enable fusion. 
Secondly, it should be able to effectively fuse the information from the aligned frames, while being robust to alignment errors. Furthermore, the lack of benchmark datasets for the general MFSR task has led to a limited interest in the MFSR problem. We address these issues by proposing a novel deep learning based approach for the MFSR problem, along with a real-world dataset.

Our network directly operates on noisy RAW bursts captured from a hand-held camera and generates a denoised, demosaicked, and super-resolved image as output. 
This is achieved by developing a novel attention-based fusion module which can adaptively merge an arbitrary number of input frames in order to produce a high quality output. 
Our approach is not limited to simple motions between the images, such as translation or homography. Instead, we  estimate dense pixel-wise optical flow to align the deep feature encoding of each input frame. The aligned representations of each frame are then merged by computing element-wise fusion weights. This allows the network to adaptively select the reliable and informative content from each image, while discarding, \eg, misaligned regions. 

The conventional approach in SISR is to train and evaluate models on synthetically generated data. However, this has been shown to not generalize to real-world images due to inaccuracies in data generation model \cite{Bulat2018ToLI,Lugmayr2019UnsupervisedLF,Lugmayr2020NTIRE2C,Lugmayr2019AIM2C}. Accurately modelling the image formation process for MFSR is further challenging due to the additional complexity introduced by camera motion. 
We therefore introduce the BurstSR dataset: the first real-world burst super-resolution dataset.
Our dataset consists of $200$ RAW bursts captured using a hand held mobile camera. Furthermore, we provide a high quality HR ground truth for each burst using a DSLR with zoom lens. We believe that our BurstSR dataset can serve as a valuable benchmark and source of training data to stimulate future research in MFSR.

\parsection{Contributions} Our main contributions are summarized as follows. \textbf{(i)} We introduce the first real world burst super-resolution dataset consisting of RAW bursts and corresponding HR ground truths. \textbf{(ii)} We propose a novel MFSR architecture which can perform joint denoising, demosaicking, and SR using bursts captured from a handheld camera. \textbf{(iii)} Our architecture employs an attention-based fusion method to adaptively merge the input images to generate high quality HR output \textbf{(iv)} We further address mis-alignment issues encountered when training on real world data by introducing a loss function which can internally correct these mis-alignments.

We perform comprehensive experiments on a synthetic dataset, as well as the BurstSR test set, in order to validate our contributions. Our approach demonstrates promising SR performance on real world bursts, significantly outperforming alternative methods in a user study. We also provide a detailed ablative study, analysing the impact of key components in the proposed MFSR architecture.

\section{Related Work}

\parsection{Single Image Super-Resolution} 
SISR is a widely studied task with a variety of proposed methods, for example based on the frequency domain~\cite{Ji2009RobustWS,Nguyen2000AWI,Rhee1999DiscreteCT}, interpolation techniques~\cite{Dai2007SoftES,Hou1978CubicSF,Li2000NewED}, sparse representations~\cite{Lu2012GeometryCS,Yang2012BilevelSC,Yang2010ImageSV} or patch and examples~\cite{Chang2004SuperresolutionTN,Freeman2002ExampleBasedS,Glasner2009SuperresolutionFA}. Dong \etal~\cite{Dong2014LearningAD} were the first to train a deep CNN to directly map the input LR image to the HR output. A number of approaches have subsequently improved upon this work using more effective network architectures~\cite{Dong2016ImageSU,Kim2016AccurateIS,Lai2017DeepLP,Lim2017EnhancedDR,Tai2017ImageSV,Zhang2018ResidualDN} and loss functions~\cite{Johnson2016PerceptualLF,Ledig2017PhotoRealisticSI,srflow,Wang2018ESRGANES,Yu2016UltraResolvingFI}. 

\parsection{Multi-Frame Super-Resolution} Compared to SISR approaches which solely rely on image priors to perform super-resolution, MFSR methods aim to merge multiple aliased  images of the same scene to reconstruct a higher resolution output. The MFSR problem was first addressed by Tsai and Huang~\cite{Tsai1984MultiframeIR}, who proposed a frequency domain based method that assumes known translations between input images. Later, Peleg \etal~\cite{Peleg1987ImprovingIR} and Irani and Peleg~\cite{Irani1991ImprovingRB} introduced the iterative back-projection approach. They estimate an initial HR image and simulate the imaging process to generate the LR images. The reconstruction error between the generated and input LR images is then minimized iteratively to refine the HR image. Hardie \etal~\cite{Hardie1998HighResolutionIR} extended this approach with an improved observation model and a regularization term. Farsui \etal~\cite{Farsiu2004MultiframeDA} proposed a joint multi-frame demosaicking and super-resolution approach using a maximum a posteriori estimation framework. Zomet~\etal~\cite{Zomet2002MultisensorS} use information from multiple sensors to perform super-resolution.
Recently, Wronski \etal~\cite{Wronski2019HandheldMS} proposed a MFSR method for hand-held cameras, where a kernel regression technique is employed to merge aligned input frames robustly. Unlike in SISR, only a few deep learning based approaches have been proposed for MFSR. Ustinova and Lempitsky~\cite{Ustinova2017DeepMF} proposed a multi-frame network for face super-resolution. Deudon \etal~\cite{Deudon2020HighResnetRF} developed HighRes-net, a MFSR network for satellite imagery. HighRes-net aligns each input frame to a reference frame implicitly, and merges them using a recursive fusion method. Another approach for satellite imagery, namely DeepSUM~\cite{Molini2020DeepSUMDN}, assumes only translation motion between frames and utilizes 3D convolution for fusion. In contrast to these previous approaches which are focused on remote sensing, we tackle the general problem of burst SR from any handheld camera.

\parsection{Learning real world super-resolution} SR approaches are commonly trained using synthetically generated LR images. However, such a training strategy has been shown to not generalize well to real-world images~\cite{Bulat2018ToLI,Lugmayr2019UnsupervisedLF,Lugmayr2020NTIRE2C,Lugmayr2019AIM2C}. A few recent works have tried to address this issue by learning real world degradation models~\cite{Bulat2018ToLI,Lugmayr2019UnsupervisedLF,deflow}. Another approach is to learn camera specific SR models directly using real world data. Such a strategy allows the network to learn the characteristics of the particular sensor, leading to improved performance~\cite{Zhang2019ZoomTL}. This is however challenging due to difficulties in collecting paired training data for SR. Zhang \etal~\cite{Zhang2019ZoomTL} address this by using LR-HR pairs captured using a zoom lens for training. In order to handle the spatial and color mis-alignments between LR-HR pairs, a novel contextual bilateral loss is employed for training. 

\begin{figure*}[t]
    \centering%
    \includegraphics[trim = 0 210 55 0, width=0.95\textwidth]{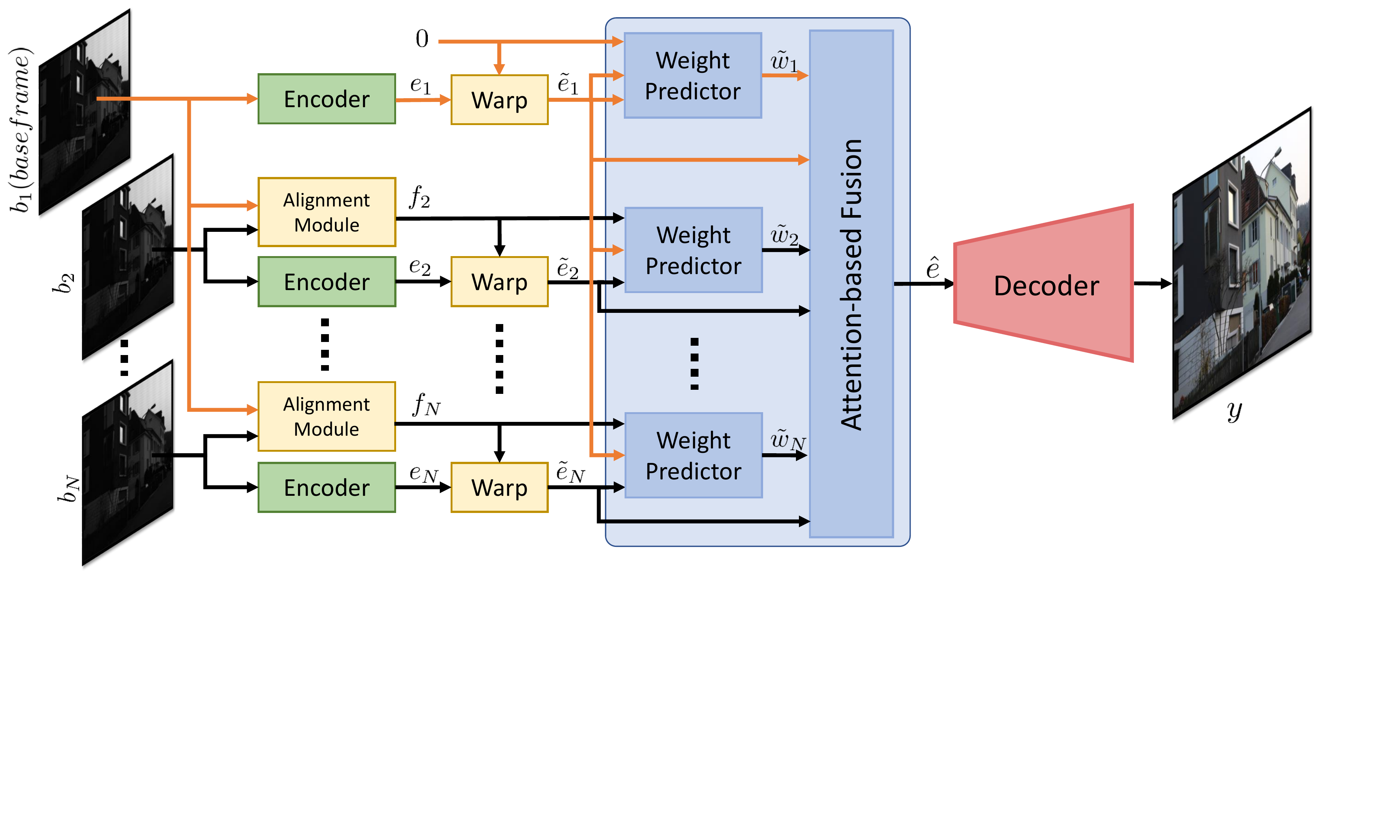}
    \caption{An overview of our burst super-resolution architecture. Each image $\bimage_i$ in the input burst is first passed independently through the encoder. The resulting feature maps are then warped to the base frame ($\bimage_1$) coordinates using the flow vectors $\flow_i$ predicted by the alignment module. The aligned feature maps are then merged using an attention-based fusion module, using fusion weights computed by the weight predictor. The merged feature map $\mergedfeat$ is then passed through to decoder module to obtain the super-resolved RGB image as output.}\vspace{-4mm}
    \label{fig:network_overview}
\end{figure*}

\section{Burst Super-Resolution Network}
In this section, we describe our burst super-resolution network. Our network inputs multiple noisy, RAW, low-resolution (LR) images captured in a single burst.
The architecture processes and combines the information in individual images to generate a high-resolution (HR) RGB image as output. Thus, our network performs joint denoising, demosaicking, and SR. Since the images in a burst are captured in a rapid sequence from a hand-held device, they include small inter-frame offsets. This ensures multiple aliased versions of the same scene, providing additional signal information for SR. Consequently, by effectively merging the information from the whole burst, our network can better reconstruct the underlying scene to generate a higher quality output, compared to single frame approaches. 

An overview of our architecture is shown in Figure~\ref{fig:network_overview}. Our network takes a RAW burst sequence $\{\bimage_i\}_{i=1}^{N}$ of any arbitrary size $N$ as input. Here, each image $\bimage_i \in \reals^{W \times H}$ is the RAW sensor data obtained from the camera. The images in the burst are first encoded independently in order to obtain deep feature representations $\{\encfeat_i\}_{i=1}^{N}$. Next, we align and warp each of the feature maps to a common reference frame $\bimage_1$ using the offsets estimated by an alignment network. The aligned feature maps are then combined by our fusion module to obtain a merged feature map $\mergedfeat$. We propose an attention-based fusion approach that predicts element-wise fusion weights. This allows the network to adaptively select the most useful information from each image in the burst. The merged feature map is then passed to the decoder module which outputs the final RGB image $y \in \reals^{sW \times sH \times 3}$, where $s$ is the super-resolution factor. We detail each network module of our architecture in the subsequent sections.

\subsection{Encoder} 
The encoder module $E$ independently maps each input burst image $\bimage_i$ to a deep feature representation $\encfeat_i$. To ensure translational invariance, we first pack each $2 \times 2$ block in the raw Bayer pattern along the channel dimension, obtaining a 4 channel image $\bimagepacked_i \in \reals^{\frac{W}{2} \times \frac{H}{2} \times 4}$ at half the initial resolution. This LR image is passed through the encoder, consisting of an initial convolutional layer followed by a series of residual blocks. In order to achieve a high-dimensional encoding that allows more effective fusion of several frames, we radically expand the feature dimensionality with a final convolutional layer. The resulting $D$-dimensional encoding $E(\bimagepacked_i) = \encfeat_i \in \reals^{\frac{W}{2} \times \frac{H}{2} \times D}$ thus achieves a rich embedding of the input image. We use $D=512$ in our experiments.

\subsection{Alignment Module} 
One of the important challenges in burst SR is that the pixel-wise displacement between the images is unknown. The displacements stem from both global camera motion and scene variations. In order to achieve an effective fusion of multiple frames, the information first needs to be aligned. We address this problem by explicitly aligning the individual image embeddings $\encfeat_i$ to a common reference LR image, called the \emph{base frame}. For convenience, we let the first image $\bimagepacked_1$ denote the base frame.
Camera motion is often modelled using a homography when imaging static and distant scenes. However, we found these assumptions to seldom hold in the real-world scenario. Thus, we allow greater flexibility in our alignment module by computing dense pixel-wise optical flow $\flow_{i}  \in \reals^{\frac{W}{2} \times \frac{H}{2} \times 2}$ between every burst image $\bimagepacked_i$ and the reference image $\bimagepacked_1$. Pixel-wise flow can capture global camera motion while also accounting for any object motion in the scene. The estimated flow vectors $\flow_{i}$ are then used to warp the feature maps $\encfeat_i$ to the base frame using a bilinear kernel
\begin{equation}
    \warpedfeat_i = \warp(\encfeat_i, \flow_{i}) \,, \quad \flow_{i} = \flowestimator(\bimagepacked_i, \bimagepacked_1)
\end{equation}
Here, $\warp$ denotes the warping operation, $\flowestimator$ is the flow estimator, while $\warpedfeat_i$ is the warped feature map. The warped feature maps $\{\warpedfeat_i\}_{i=1}^N$, as well as the computed flow vectors $\{\flow_{i}\}_{i=1}^N$ are then passed to the fusion module. Here, the flow vectors $\flow_{1}$ for the base frame is set to 0.
While any state-of-the-art optical flow network~\cite{IMKDB17,Sun2018PWCNetCF,Teed2020RAFTRA,GOCor_Truong_2020} can be employed as our flow estimator $\flowestimator$, we use the PWC-Net~\cite{Sun2018PWCNetCF} approach due to it's high accuracy and speed. 
Since PWC-Net is trained to operate on RGB images, we discard one of the two green channels in $\bimagepacked_i$ to generate input RGB images. 

\subsection{Fusion Module} 
The fusion module combines information across the individual burst images to generate a merged feature embedding $\mergedfeat$. In order to be able to operate on bursts of arbitrary sizes, the fusion module must be able to merge any number of input frames. Consequently, it is infeasible to \eg directly concatenate the input feature maps along the channel dimension. We further found simple pooling operations such as element-wise max or average pool across the burst to provide unsatisfactory results. This is because the fusion module needs be able to merge adaptively based on \eg image content, noise levels, etc. For instance, it can be beneficial to have uniform fusion weights for textureless regions in order to perform denoising. On the other hand, it is preferable to have low fusion weights for any mis-aligned frame in order to avoid ghosting artifacts.  
We therefore propose an attention-based fusion approach, where element-wise fusion weights are predicted by a weight predictor network $\weightpred$. This provides flexibility to the network to effectively extract the useful information from each image, while also being able to process arbitrary number of input images.

The weight predictor network $\weightpred$ utilizes both the aligned feature maps $\warpedfeat_i$ and the flow vectors $\flow_{i}$ to estimate the unnormalized attention weights $\fusionweightraw_i \in \reals^{\frac{W}{2} \times \frac{H}{2} \times D}$ for each embedding $\warpedfeat_i$. We first project $\warpedfeat_i$ to a lower dimension feature map $\warpedfeatproj_i$ for computational efficiency. To compute the attention weights for $\warpedfeat_i$, we use the projected base frame feature map $\warpedfeatproj_1$, as well as the residual $r_{i} = \warpedfeatproj_i - \warpedfeatproj_1$ between $\warpedfeatproj_i$ and $\warpedfeatproj_1$. The base frame map $\warpedfeatproj_1$ contains information about the local image content. This is informative to determine \eg whether to use uniform fusion weights to achieve denoising, or perform edge-aware fusion in order to avoid over smoothing edges. On the other hand, the residual $r_{i}$ can provide an estimate of alignment errors and thus help assign low fusion weights to misaligned regions. Additionally, we use the flow vectors $\flow_{i}$ for weight estimation as they provide the sub-pixel sampling location of the image data. 
We obtain the sub-pixel offset by computing modulo $1$ of the flow vectors $\flow_{i}$ and pass it through a small CNN to obtain the flow features $\flowfeatures_{i}$. The reference frame features $\warpedfeatproj_1$, the feature residual $r_{i}$, and the flow features $\flowfeatures_{i}$ are concatenated along the channel dimension and passed through a residual network to obtain the raw fusion weights $\fusionweightraw_i$. The raw fusion weights are then normalized across the burst using a softmax function to obtain the final attention weights $\fusionweight_i$. The merged feature map $\mergedfeat$ is then be obtained as the following weighted sum,
\begin{equation}
\mergedfeat = \sum_{i=1}^{N} \fusionweight_i \cdot \encfeat_i \,, \;\; \fusionweight_i = \frac{e^{\fusionweightraw_i}}{\sum_j e^{\fusionweightraw_j}} ,\;\; \fusionweightraw_i = \weightpred\!(\warpedfeat_1, r_{i}, \flowfeatures_{i}).
\end{equation}
Here, $\cdot$ denotes element-wise multiplication. The merged feature map $\mergedfeat$ is then passed to the decoder module to generate the final output.

\subsection{Decoder}
The decoder module generates the output high-resolution RGB image from the fused feature map $\mergedfeat$. We first project the input feature map to $128$ channels and pass it through a residual network. Next, we upsample this to the desired resolution $sH \times sW$ using sub-pixel convolution~\cite{Shi2016RealTimeSI}. We use a convolution layer to increase the feature dimension to $2^2s^2D'$, obtaining a tensor of shape $\frac{H}{2} \times \frac{W}{2} \times 2^2s^2D'$. The feature vectors at each spatial location are then re-arranged into a $2s \times 2s \times D'$ map to obtain a higher resolution feature map of shape $H \times W \times D'$. Here, $D'$ is the output feature dimension of the sub-pixel convolution layer.
Compared to performing na\"ive upsampling using \eg bilinear interpolation, sub-pixel convolution allows us to effectively decode the sub-pixel information encoded in the different feature channels. 
In order to avoid checkerboard artifacts, we use the ICNR initialization~\cite{Aitken2017CheckerboardAF} for the sub-pixel convolution layer and additionally apply Gaussian smoothing to its output. 
The upsampled feature map is then passed through another set of residual blocks, followed by a conv layer to obtain the high resolution RGB image $y$.

\section{BurstSR Dataset}
The aim of this work is to propose a burst SR method for real-world photography applications. 
In order to validate the performance of our approach, it is essential to train and evaluate our models on real data. 
Hence, we collect a new dataset, called BurstSR. To the best of our knowledge, it is the first real world burst super-resolution dataset.
The BurstSR dataset consists of $200$ RAW burst sequences, and corresponding high-resolution ground truths. Each burst sequence contains $14$ RAW images captured using identical camera settings (\eg exposure, ISO). All bursts are captured using a handheld smartphone camera. Our dataset therefore contains natural hand tremors, resulting in small random offsets between the images within a burst that are essential for MFSR~\cite{Tsai1984MultiframeIR}.
For each burst sequence, we also capture a high-resolution image using a DSLR camera mounted on a tripod to serve as ground truth. 
Our BurstSR dataset will be released upon publication. We believe that it can serve as an important training set and benchmark for the community, in order to raise the interest in the important MFSR problem.

We capture the burst images in our datset using a handheld Samsung Galaxy S8 smartphone camera. 
In order to capture and store RAW bursts, we developed a custom app using Camera2 API. On pressing the shutter, the app runs the camera's auto-focus, auto-exposure, and auto-white-balance algorithms to determine the camera settings. These settings are then used to capture a fixed number of RAW images. 
The corresponding ground truth images for each burst are collected using a Canon 5D Mark IV DSLR camera mounted on a tripod. We use a zoom lens with a focal length of 70mm to obtain images with $\approx4$ times higher spatial resolution compared to burst images captured from the phone camera. The images are taken using a smaller aperture size (F18) to have a wider depth of field. Other capture settings are automatically determined by the camera. We hold the phone camera just above the DSLR when taking bursts in order to minimize misalignments between the two images. Additionally, we use a timer on the DSLR to synchronize the capture time between the two cameras. In order to minimize the effect of any error in temporal synchronization, we try to capture static scenes with little (e.g.\ leaves moving due to wind) or no motion. 
We collect $200$ bursts in total, which are split into train, validation, and test sets consisting of $160$, $20$, and $20$ sequences, respectively.

\section{Training}
In this section, we describe our training pipeline in detail. Due to the high cost and effort associated with collecting real-world paired data for MFSR, it is impractical to obtain large scale real world datasets for training our model from scratch. 
We therefore exploit methods for synthetic data generation to first pre-train our networks. The resulting model serves as a strong initialization, which is then finetuned on our BurstSR dataset to perform real-world SR. 

\subsection{Synthetic data training}
\label{sec:synthetic_training}
We generate synthetic RAW bursts for pre-training our model using the sRGB images from the training split of Zurich RAW to RGB dataset~\cite{ignatov2020replacing}. Given a sRGB image, we apply the inverse camera pipeline described in~\cite{Brooks2019UnprocessingIF} to obtain raw sensor values. 
Next, we generate a synthetic burst of size $N$ by applying random translations and rotations to the converted RGB image. The translation and rotation values are sampled independently from the range [-24, 24] pixels and [-1, 1] degrees, respectively. The transformed images are then downsampled by the desired super-resolution factor $s$ to obtain the low resolution RGB burst. We use bilinear kernel for both image translation/rotation and downsampling. Next, we add shot and read noise to the burst images, as described in~\cite{Brooks2019UnprocessingIF}.  We then discard two color channels per pixel according to the Bayer CFA to obtain the mosaicked RAW burst. We extract $96 \times 96$ crops from the resulting RAW burst for our training. Our network is trained in a fully supervised manner by minimizing the $L_1$ loss between the network prediction and the ground truth image. The loss is computed in the linear sensor space, before any post processing \eg gamma compression or tone-mapping.

\subsection{Real data training}
\label{sec:real_data_training}
In order to reconstruct the HR image using multiple aliased LR observations, a MFSR model needs to learn the image formation process in a camera. 
However, due to differences in the image formation process in a real camera and the one modelled by our synthetic pipeline, a network trained on only synthetic data is thus expected to have sub-optimal performance when applied to real data. Hence, we fine-tune the pre-trained synthetic data model on our real world BurstSR dataset in order to adapt the model to the particular camera sensor. 

\parsection{Data Processing}
Here, we describe the pipeline used to pre-process the collected BurstSR data for training. Since the images captured using phone and DSLR cameras have different field of views (FOV), we first crop out matching field of view from each image in the burst. This is done by estimating a homography between the first image in the burst and DSLR image using SIFT~\cite{Lowe1999ObjectRF} and RANSAC~\cite{Fischler1981RandomSC}. Next, we extract $160 \times 160$ crops from the burst images in a sliding window manner, with a stride of $80$ pixels. For each crop, we again estimate homography between the crop and the corresponding region in DSLR image to perform local alignment. The aligned DSLR image region is then downsampled to $160s \times 160s$ to obtain the ground truth crop. In order to filter out crops with incorrect alignment, we discard phone-DSLR pairs which have a normalized cross correlation of less than $0.9$ between them. 

\parsection{Training loss} There are several challenges when training our model on real bursts due to the unavoidable mis-alignments between the input burst and the ground truth. Firstly, even though we align the burst images to DSLR using homography, there can still be misalignments between the pair due to perspective shifts, error in homography estimation, etc. Secondly, since the burst and ground truth images are captured using two different sensors, there is a color mis-match between the two. Thus, it is not feasible to train the model by directly computing a pixel-wise error between the network prediction $y$ and the ground truth $\gtimage$. 

In order to handle the spatial mis-alignment issue, we first estimate the optical flow $f_\text{Pred,GT}$ between the prediction and ground truth using PWC-Net. The estimated flow is then used to warp the network prediction to the ground truth co-ordinates. Next, we estimate a global color mapping between the burst and the ground truth in order to handle the color mis-match. We first downsample the ground truth image to the same resolution as the input burst images. The estimated flow $f_\text{Pred,GT}$ is then used to align the first image in the burst to the downsampled ground truth. In order to minimize the effect of small mis-alignments, we apply Gaussian smoothing on both the images to obtain the processed burst image $\bar{\bimage}_1$ and ground truth image $\bar{y}_\text{GT}$. Given this aligned input-ground truth pair, we estimate a pixel-wise color mapping $\ccm$ between the two images. We assume that the color mapping is linear and model it as a $3\times3$ color correction matrix, which is computed by minimizing a least squares loss. Using the estimated color correction matrix, we can map the network prediction to the same color space as the ground truth and compute pixel-wise error.
Our training loss $\ell(y, \gtimage)$ is thus computed as
\begin{equation}
	\ell(y, \gtimage) = \sum_{n}m^n \cdot L_{1}(\hat{y}^n, y_\text{GT}^n) \,, \quad \hat{y} = \ccm(\warp(y, f_\text{Pred,GT}))
\end{equation}
Here, $\hat{y}$ is the aligned and color mapped network prediction. The summation is over all pixel coordinates $n$ in the image. The factor $m^n$ is a binary masking variable used to filter out image regions which are not aligned correctly. It is set to $0$ in regions where the error $R = \left\lVert\bar{y}_\text{GT} - \ccm(\bar{\bimage}_1)\right\rVert_2$ after color mapping the processed burst image $\bar{\bimage}_1$ is greater than a threshold. Note that the images $\bar{y}_\text{GT}$ and $\bar{\bimage}_1$ have lower-resolution compared to the model prediction $y$. Thus the error map $R$ is upsampled to the same resolution as $y$, before computing the mask $m$.

\subsection{Training details}
We use pre-trained PWC-Net weights for our flow estimator $F$. All other modules are initialized using~\cite{He2015DelvingDI}. Our model is first trained using the synthetic data for 100k iterations, and then fine-tuned on the BurstSR dataset for an additional 40k iterations. We use the ADAM~\cite{Kingma2015AdamAM} optimizer for out training. Data augmentation is performed using random cropping and flipping. Our entire training takes 30 hours on a single Nvidia V100 GPU. All our networks are trained using a burst size of 8.

\section{Experiments}
We perform comprehensive qualitative and quantitative evaluation of our approach. All our experiments are performed for super-resolution by a factor $s=4$. Additional details and results are provided in the suppl.\ material.

\subsection{Analysis of our approach}
\label{sec:ablation}
Here, we analyze the impact of different components in the proposed burst SR architecture. We report results on a synthetically generated test set containing 300 bursts, as well as our BurstSR validation dataset. The synthetic test set is generated using the pipeline described in Sec~\ref{sec:synthetic_training}, using sRGB images from the test set of the Zurich RAW to RGB dataset~\cite{ignatov2020replacing}. 
We evaluate the networks trained using only the synthetic training data on this set. 
Since an accurate ground truth HR image is naturally available, the synthetic test set allows us to evaluate the impact of different architectural choices. We also report results on our BurstSR validation set, using the models fine-tuned on BurstSR training set. Since the input burst and HR ground truth in BurstSR are captured using different cameras, there exists spatial and color misalignments between them. We therefore align the network prediction to the ground truth and perform color transformation as described in Sec~\ref{sec:real_data_training}. The resulting image is then compared with the ground truth in order to compute performance metrics. We report the standard fidelity based metrics PSNR and SSIM~\cite{Wang2004ImageQA}, as well as the learned perceptual score LPIPS~\cite{Zhang2018TheUE} on both datasets. All metrics are computed in linear sensor space. Note that the images in our BurstSR dataset are generally underexposed, leading to high PSNR scores for all methods. 
Unless specified, all the methods are evaluated using a burst size of 8.

\parsection{Impact of using multiple frames} Here, we investigate the impact of using multiple frames for SR by comparing our MFSR approach with a single image baseline. We train a SISR network with exactly the same encoder and decoder architecture as employed in our approach. In order to ensure that the SISR performance is not limited by model capacity, we increased the depth of the single image network until its performance saturated. We compare this single image baseline with our multi-frame approach, evaluated using bursts of different sizes. The result of this comparison is shown in Table~\ref{tab:single_frame}. 
Even when using only 4 input frames, our approach significantly outperforms the single image baseline with an improvement of $0.76$ dB in PSNR on the synthetic set. Note that although our model is trained using a fixed burst size of 8, it generalizes to bursts with varying input sizes, providing a consistent improvement with increasing burst size. This shows that our approach can effectively utilize the information from multiple frames in order to improve SR performance. When using bursts of size 14, our approach obtains an improvement of $2.67$ dB in PSNR on the synthetic set, clearly demonstrating the advantages of using multiple frames for SR.

\begin{table}[!t]
	\centering\vspace{-1mm}
	\resizebox{\columnwidth}{!}{%
		\begin{tabular}{l|ccc|ccc}
\toprule
&\multicolumn{3}{c|}{Synthetic data} & \multicolumn{3}{c}{BurstSR} \\
&PSNR $\uparrow$&LPIPS $\downarrow$&SSIM $\uparrow$&PSNR $\uparrow$&LPIPS $\downarrow$&SSIM $\uparrow$\\\midrule
Single Image& 36.42 & 0.123 & 0.913 & 46.41 & 0.041 & 0.979 \\
Burst-2& 34.90 & 0.133 & 0.893 & 46.10 & 0.040 & 0.977 \\
Burst-4& 37.18 & 0.092 & 0.927 & 47.06 & 0.033 & 0.981 \\
Burst-8& 38.61 & \textbf{0.084} & 0.941 & 47.52 & 0.031 & 0.983 \\
Burst-14 & \textbf{39.09} & \textbf{0.084} & \textbf{0.945} & \textbf{47.76} & \textbf{0.030} & \textbf{0.984}\\\bottomrule
\end{tabular}

	}\vspace{1mm}%
	\caption{Comparison of the baseline SISR network with our multi-frame approach, evaluated using different number of input frames. 
	}
	\label{tab:single_frame}%
	\vspace{-1mm}
\end{table}

\begin{table}[!t]
	\centering\vspace{-1mm}
	\resizebox{\columnwidth}{!}{%
		\begin{tabular}{l|ccc|ccc}
\toprule
&\multicolumn{3}{c|}{Synthetic data} & \multicolumn{3}{c}{BurstSR} \\
&PSNR $\uparrow$&LPIPS $\downarrow$&SSIM $\uparrow$&PSNR $\uparrow$&LPIPS $\downarrow$&SSIM $\uparrow$\\\midrule
Ours& \textbf{38.61} & \textbf{0.084} & \textbf{0.941}  & \textbf{47.52} & \textbf{0.031} & \textbf{0.983} \\
No Alignment& 36.66 & 0.119 & 0.915 & 46.50 & 0.040 & 0.979 \\
Single Image& 36.42 & 0.123 & 0.913  & 46.41 & 0.041 & 0.979\\\bottomrule
\end{tabular}

	}\vspace{1mm}%
	\caption{Comparison of our approach performing explicit alignment with a baseline which does not employ an alignment module.}
	\label{tab:alignment}%
	\vspace{-3mm}
\end{table}
\parsection{Impact of alignment module} We analyse the impact of the alignment module in our architecture by evaluating a baseline network which does not perform any explicit alignment. We directly concatenate the encoded \textit{base} frame features to all other frames, and pass the resulting feature maps through additional residual blocks, before merging them. The result of this comparison is shown in Table~\ref{tab:alignment}. Our approach, performing explicit sub-pixel alignment using a flow estimator, outperforms the baseline \textbf{No Alignment} with an improvement of $1.02$ dB in PSNR on the BurstSR validation set. 
Interestingly, the No Alignment network only obtains a slight improvement over the SISR baseline. These results show that accurate alignment of input frames is essential in order to benefit from multiple frames.

\parsection{Analysis of fusion architecture} We compare our proposed attention-based fusion module with 4 different alternatives. i) \textbf{MaxPool}: The encoded feature maps are merged by performing element-wise max pooling across the burst. ii) \textbf{AvgPool}: The merged feature map is computed as element-wise mean across the burst. iii) \textbf{Concatenate}: The encoded feature maps are concatenated along the channel dimension to obtain the merged features. Note that this architecture is constrained to operate on bursts of fixed size.
iv) \textbf{RecMerge}: The recursive fusion strategy proposed in~\cite{Deudon2020HighResnetRF}. Pairs of encoded feature maps are concatenated and passed through a small network to merge them. This process is repeated recursively until a single merged feature map is obtained. All four baseline networks employ the same encoder, decoder, and alignment modules as used in our approach to ensure a fair comparison.

The result of this analysis is shown in Table~\ref{tab:fusion}. We observe that MaxPool and AvgPool approaches obtain poor results, indicating that simple pooling operations are insufficient to perform effective merging. Both Concatenate and RecMerge achieve better results with PSNR of $37.80$ dB and $37.55$ dB respectively, on the synthetic set. Our attention-based fusion obtains the best results on both the synthetic set as well as BurstSR, showing that it can effectively merge the information from the input frames. 

\begin{table}[!t]
	\centering\vspace{-1mm}
	\resizebox{\columnwidth}{!}{%
		\begin{tabular}{l|ccc|ccc}
\toprule
&\multicolumn{3}{c|}{Synthetic data} & \multicolumn{3}{c}{BurstSR} \\
&PSNR $\uparrow$&LPIPS $\downarrow$&SSIM $\uparrow$&PSNR $\uparrow$&LPIPS $\downarrow$&SSIM $\uparrow$\\\midrule
Ours& \textbf{38.61} & \textbf{0.084} & \textbf{0.941}  & \textbf{47.52} & \textbf{0.031} & \textbf{0.983}\\
MaxPool& 36.24 & 0.116 & 0.912 & 46.74 & 0.039 & 0.980 \\
AvgPool& 35.45 & 0.131 & 0.902 & 46.53 & 0.040 & 0.979 \\
Concatenate& 37.80 & 0.098 & 0.928 & 47.17 & 0.034 & 0.981 \\
RecMerge& 37.55& 0.098 & 0.927 & 47.12 & 0.033 & 0.981 \\\bottomrule
\end{tabular}

	}\vspace{1mm}%
	\caption{Analysis of different fusion approaches for merging the information from input frames. 
	}
	\label{tab:fusion}%
	\vspace{-2mm}
\end{table}

\begin{table}[!t]
	\centering\vspace{-1mm}
	\resizebox{\columnwidth}{!}{%
		\begin{tabular}{l|ccc|ccc}
\toprule
&\multicolumn{3}{c|}{Synthetic data} & \multicolumn{3}{c}{BurstSR} \\
&PSNR $\uparrow$&LPIPS $\downarrow$&SSIM $\uparrow$&PSNR $\uparrow$&LPIPS $\downarrow$&SSIM $\uparrow$\\\midrule
Only Feature& 37.46 & 0.101 & 0.927  & 47.11 & 0.034 & 0.981 \\
Only Residual& 38.14 & 0.093 & 0.935  & 47.46 & 0.031 & 0.982\\
Residal+Base & 38.41 & 0.085 & 0.939  & 47.46 & \textbf{0.030} & \textbf{0.983}\\
Residal+Base+Flow& \textbf{38.61} & \textbf{0.084} & \textbf{0.941} & \textbf{47.52} & 0.031 & \textbf{0.983}\\
\bottomrule
\end{tabular}

	}\vspace{1mm}%
	\caption{Impact of different inputs used by the weight predictor. 
	}
	\label{tab:weight_predictor}%
	\vspace{-3mm}
\end{table}
\parsection{Analysis of weight predictor network} Here, we analyse the impact of different inputs used by our weight predictor network to determine the element-wise fusion weights. We evaluate 4 different versions of the weight predictor, using different sets of inputs, i) \textbf{Only Feature}: Only the projected feature map $\warpedfeatproj_i$ is used. ii) \textbf{Only Residual}: Only the feature residual $r_{i} = \warpedfeatproj_i - \warpedfeatproj_1$ is used. iii) \textbf{Residual+Base}: Both the feature residual $r_{i}$ and the base frame features $\warpedfeatproj_1$ are used. iv) \textbf{Residual+Base+Flow}: The feature residual $r_{i}$, base frame features $\warpedfeatproj_1$, as well as the flow features $\flowfeatures_{i}$ are used. The result of this comparison is shown in Table~\ref{tab:weight_predictor}. Compared to using only the input feature $\warpedfeatproj_i$, using the residuals $r_{i}$ instead leads to better performance. Additionally using the base frame features $\warpedfeatproj_1$ improves the performance further by $0.27$ dB in PSNR on the synthetic set. The best results are obtained when using the feature residual $r_{i}$, the base frame features $\warpedfeatproj_1$, and the flow features $\flowfeatures_{i}$ together, showing that they each provide complementary information to the weight predictor.

\begin{figure*}[t]
    \centering%
    \includegraphics[trim = 0 0 0 0, width=\textwidth]{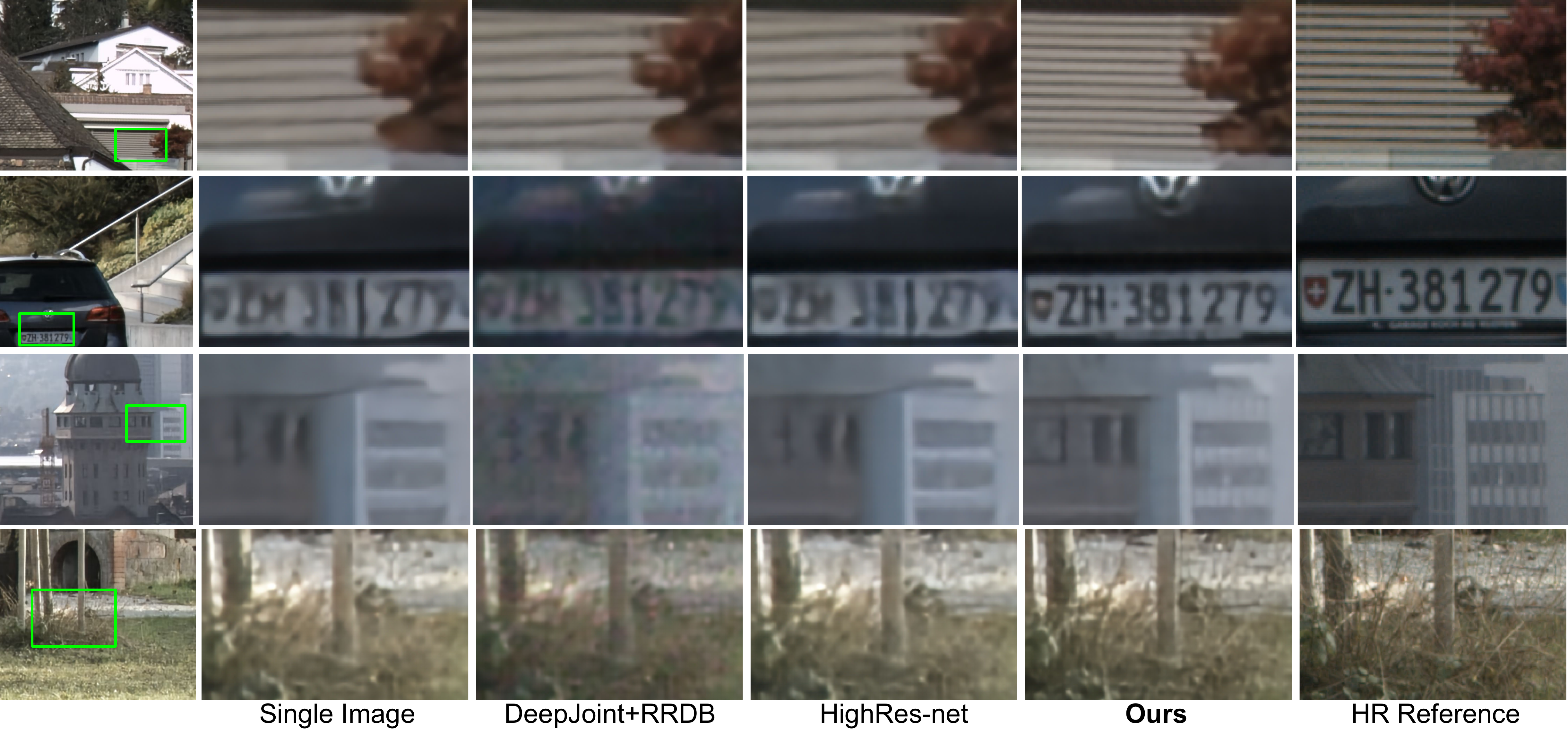}\vspace{-1mm}
    \caption{Qualitative comparison of our approach on real world bursts from the BurstSR test set. Our approach can effectively merge information from multiple frames to reconstruct high-frequency image details.}\vspace{-3mm}
    \label{fig:qual_fig}
\end{figure*}

\subsection{Comparison with other approaches}
In this section, we evaluate our proposed burst super-resolution network on the test set of our BurstSR dataset. We compare our approach with three methods: i) \textbf{Single Image} Our SISR baseline network; ii) \textbf{DeepJoint+RRDB} A two stage network which performs single frame demosaicking and denoising using DeepJoint~\cite{Gharbi2016DeepJD} and super-resolves the resulting RGB image using the RRDB~\cite{Wang2018ESRGANES} network; and iii) \textbf{HighRes-net} A recent deep learning based MFSR approach~\cite{Deudon2020HighResnetRF} proposed for remote sensing applications. HighRes-net performs implicit registration of the input frames, without using any independent alignment module. Fusion is performed in a recursive manner. We use pre-trained weights for the DeepJoint and RRDB networks. The Single Image baseline, as well as HighRes-net, are trained to perform joint denoising, demosiacking, and SR using the exact training pipeline used by our approach. In order to ensure a fair comparison, we increased the depth of the original HighRes-net network to have the same number of residual blocks as in our approach.  

\begin{table}[!t]
	\centering\vspace{-1mm}
	\resizebox{\columnwidth}{!}{%
		\begin{tabular}{l|cc|ccc}
\toprule
&MOR $\downarrow$&\%\text{Top} $\uparrow$&PSNR $\uparrow$&LPIPS $\downarrow$&SSIM $\uparrow$\\\midrule
DeepJoint+RRDB&3.42&8.9& 42.13 & 0.088 & 0.957 \\
Single Image&2.41&18.3& 44.02 &  0.051 & 0.972 \\
HighRes-net&2.36&19.2& 43.99 &  0.051 & 0.972\\
\textbf{Ours}&\textbf{1.81}&\textbf{53.6}& \textbf{45.17} &  \textbf{0.037} & \textbf{0.978}\\
\bottomrule
\end{tabular}

	}\vspace{1mm}%
	\caption{Comparison of our method with existing SR approaches on the BurstSR test set. We report the results of our user study, as well as the standard quality metrics PSNR, LPIPS, and SSIM. }
	\label{tab:user_study}%
	\vspace{-3mm}
\end{table}

We conducted a user study on Amazon Mechanical Turk to compare the four approaches. We obtain the HR prediction for each of our network on the $20$ test images. Next, we extract 15 random $200 \times 200$ crops from each of our $20$ test images. Each of the 300 crops are then resized to $400 \times 400$ using nearest neighbor interpolation. We show the participants the ground truth HR image, as well as the network predictions. The participants are asked to rank the predictions from the 4 approaches according to the visual quality \wrt the provided DSLR reference image. The network predictions were anonymized and randomized in order to avoid any bias. We obtained $5$ independent rankings for each crop. The mean ranking (MOR) over all the crops, as well as the percentage of times a method was ranked first ($\%\text{Top}$) are shown in Table~\ref{tab:user_study}. Our approach obtains a MOR of $1.81$, significantly better than all other approaches. Furthermore, our approach is ranked as the best among all methods $53.6\%$ of the times, more than $2.5$ times the second best method.  We also report the PSNR, LPIPS, and SSIM scores on the test set, computed as described in Sec.~\ref{sec:ablation}. A qualitative comparison is also provided in Fig.~\ref{fig:qual_fig}. Our approach obtains the best results in terms of all three metrics, outperforming HighRes-net by $1.18$ dB in terms of PSNR.

\section{Conclusions}
We address the problem of real-world multi-frame super-resolution. We introduce a new dataset BurstSR containing RAW burst sequences captured from a handheld camera, and corresponding high-resolution ground truths obtained using a zoom lens. We further propose a multi-frame super-resolution network which can adaptively combine the information from multiple input images using an attention-based fusion. Our approach obtains promising results on real world bursts, outperforming both single frame as well as multi-frame alternatives.

\noindent\textbf{Acknowledgments}: This work was supported by a Huawei Technologies Oy (Finland) project, the ETH Z\"urich Fund (OK), an Amazon AWS grant, and an Nvidia hardware grant.

{\small
\bibliographystyle{ieee_fullname}
\bibliography{references}
}

\clearpage
\setcounter{section}{0}
\renewcommand{\thesection}{\Alph{section}}

\begin{center}
	\textbf{\large Supplementary Material}
\end{center}

We provide additional details and analysis of our approach in this supplementary material. In Section~\ref{sec:net_arch}, we provide additional details about our network architecture. We analyse the impact of sub-pixel shifts in the input images for MFSR in Section~\ref{sec:impact_shift}, while the impact of training dataset for real world SR is analysed in Section~\ref{sec:impact_dataset}.
Section~\ref{sec:loss} provides a qualitative analysis of the impact of our training loss (3) used to train our networks on the BurstSR dataset. Additional qualitative comparison with existing super-resolution approaches are provided in Section~\ref{sec:qual}

\section{Network Architecture}
\label{sec:net_arch}
Here, we provide additional details about our burst super-resolution network architecture.

\begin{figure}[t]
	\centering%
	\includegraphics[trim = 0 0 0 0, width=\columnwidth]{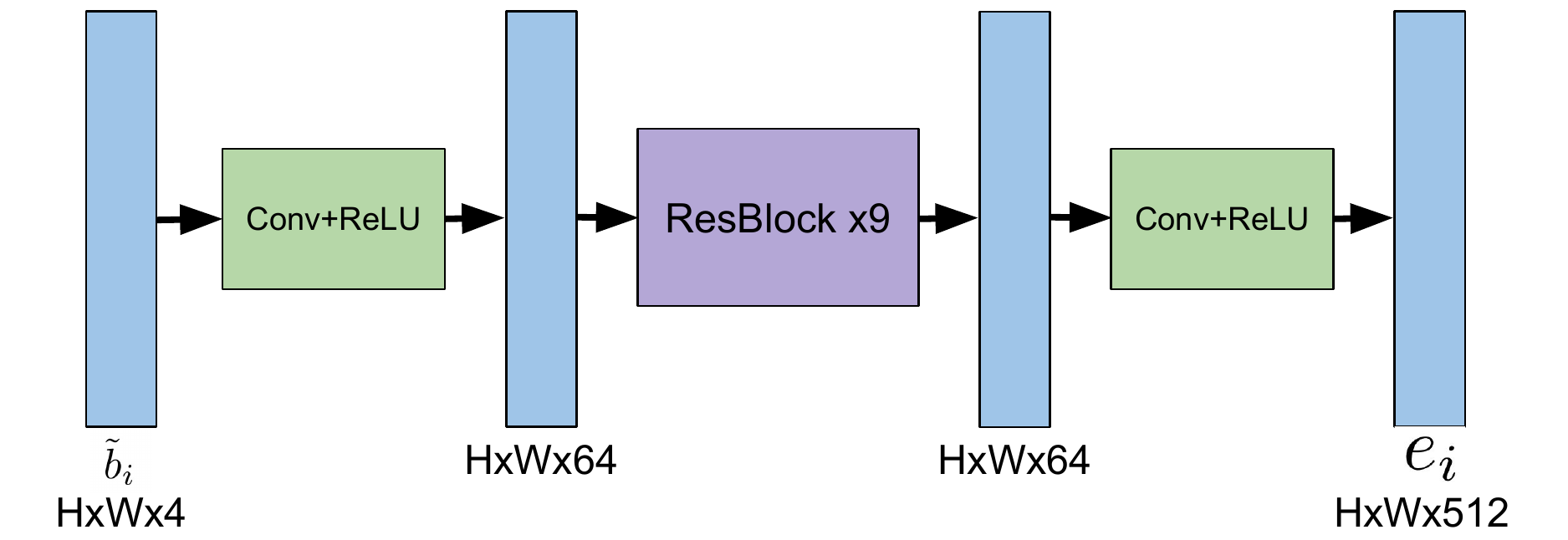}
	\caption{The network architecture employed for the Encoder module $E$.}\vspace{0mm}
	\label{fig:encoder}
\end{figure}

\parsection{Encoder} The encoder module maps the packed RAW image $\bimagepacked_i$ to a $64$ dimensional feature embedding using a convolution layer. The resulting feature map is processed by 9 residual blocks, before being passed to another convolution layer which expands the feature dimensionality to $512$. An illustration of the Encoder module is provided in Figure \ref{fig:encoder}.

\begin{figure*}[t]
	\centering%
	\includegraphics[trim = 0 0 0 0, width=\textwidth]{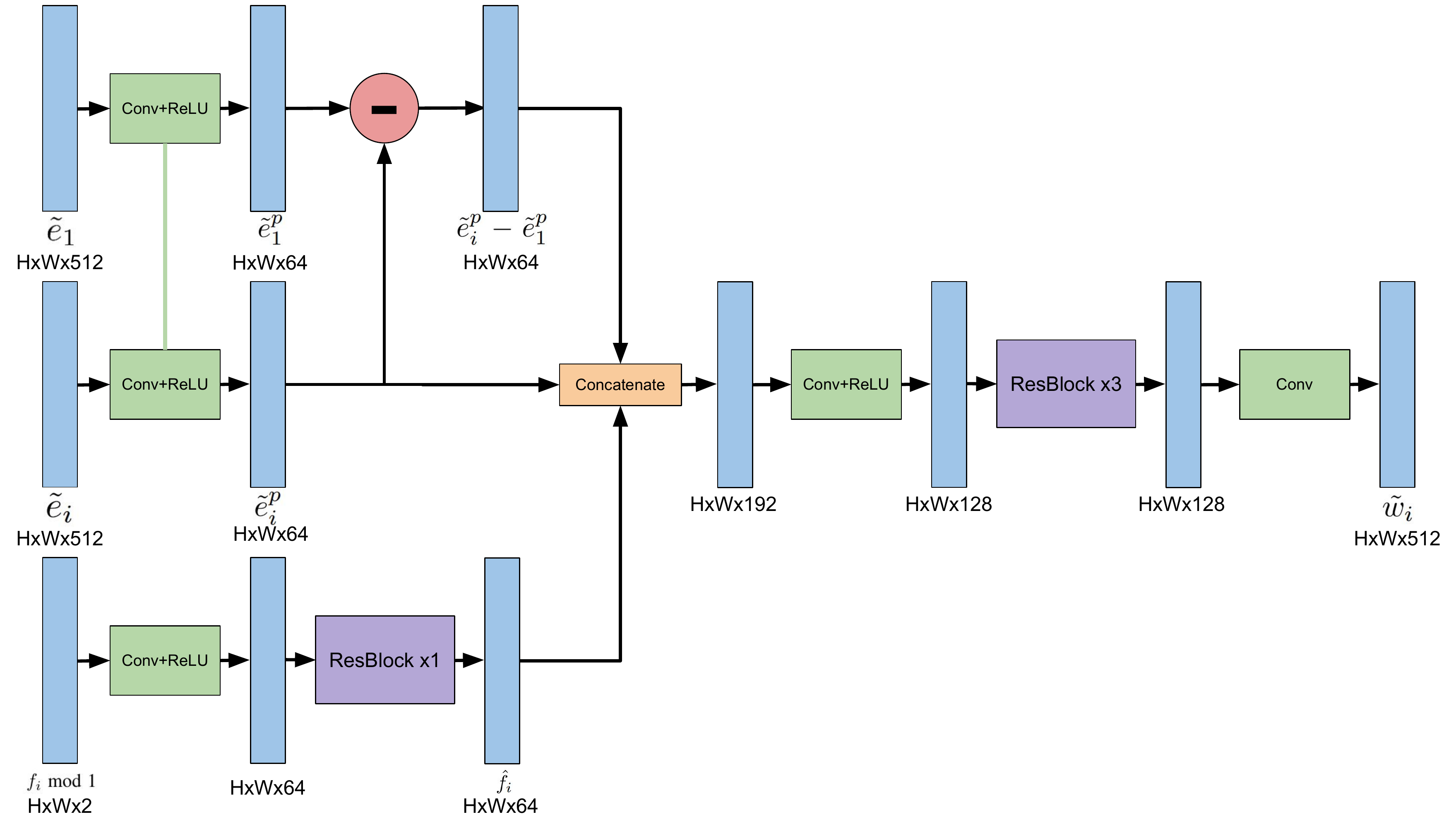}
	\caption{The network architecture employed for the Weight Predictor module $W$.}\vspace{0mm}
	\label{fig:weightpred}
\end{figure*}

\parsection{Weight Predictor} The weight predictor module computes the un-normalized element-wise fusion weights for each aligned feature embedding $\warpedfeat_i$. It first projects the feature embeddings $\warpedfeat_i$ and $\warpedfeat_1$ to 64 dimensional feature maps $\warpedfeatproj_i$ and $\warpedfeatproj_1$ respectively, using a convolution layer with shared weights. Additionally, the weight predictor module also extracts flow features $\flowfeatures_i$ using the flow vectors $\flow_i$. The modulo 1 of the flow vectors, $\flow_i$ mod 1, is first passed through a convolution layer, followed by a residual block to obtain $64$ dimensional flow features $\flowfeatures_i$. The flow features $\flowfeatures_i$, the projected feature embedding $\warpedfeatproj_i$, and the residual $\warpedfeatproj_i - \warpedfeatproj_1$ are then concatenated along the channel dimension, and passed through a convolution layer. The output 128 dimensional feature map is processed by 3 residual blocks, before being passed to a final convolution layer which predicts raw element-wise fusion weights $\fusionweightraw$.  An illustration of the weight predictor module is provided in Figure \ref{fig:weightpred}.
 
\begin{figure*}[t]
	\centering%
	\includegraphics[trim = 0 0 0 0, width=\textwidth]{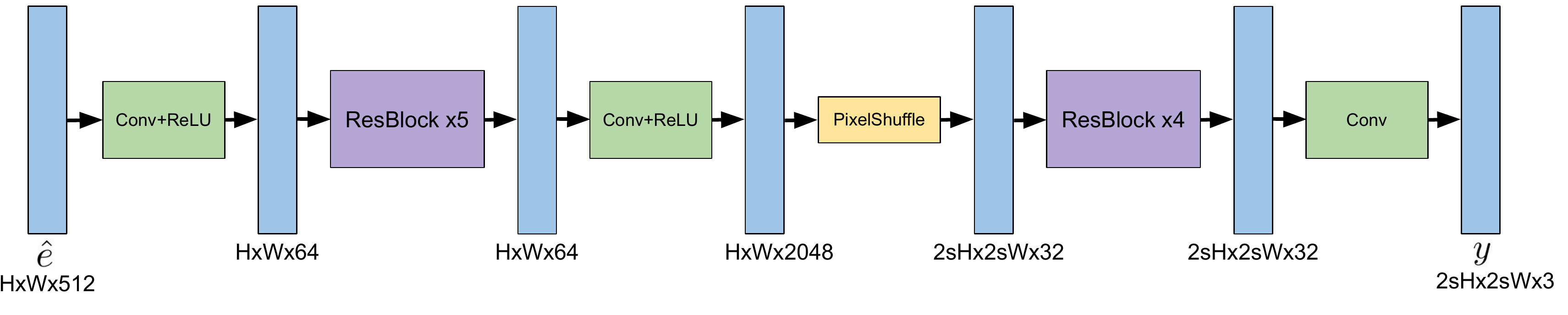}
	\caption{The network architecture employed for the Decoder module $D$.}\vspace{0mm}
	\label{fig:decoder}
\end{figure*}

\parsection{Decoder} The decoder module projects the merged feature map $\mergedfeat$ to a 64 dimensional feature space. The projected features are then passed through 5 residual blocks, before being passed to the sub-pixel convolution layer, which upsamples the feature map by a factor $2s$. The sub-pixel convolution layer first increases the feature dimensionality  to $2^2s^232$ using a convolution layer. The feature vectors at each spatial location are then re-arranged into a $2s \times 2s \times 32$ map to obtain a 32 dimensional feature map with $2s$ times higher resolution compared to the input. The upsampled feature map is then processed by 4 residual blocks, before being passed to a convolution layer which predicts the output RGB image. An illustration of the Decoder module is provided in Figure \ref{fig:decoder}.

\begin{table}[!t]
	\centering\vspace{-1mm}
	\resizebox{0.7\columnwidth}{!}{%
		\begin{tabular}{lccc}
\toprule
&PSNR $\uparrow$&LPIPS $\downarrow$&SSIM $\uparrow$\\\midrule
Ours& \textbf{38.61} & \textbf{0.084} & \textbf{0.941} \\
No Shifts& 37.00 & 0.106 & 0.920 \\
Single Image& 36.42 & 0.123 & 0.913 \\
\bottomrule
\end{tabular}

	}\vspace{1mm}%
	\caption{Impact of sub-pixel shifts in the input burst for MFSR. Results are shown on the synthetic test set.}
	\label{tab:noshift}%
	\vspace{-2mm}
\end{table}

\begin{table}[!t]
	\centering\vspace{-1mm}
	\resizebox{0.7\columnwidth}{!}{%
		\begin{tabular}{lccc}
\toprule
&PSNR $\uparrow$&LPIPS $\downarrow$&SSIM $\uparrow$\\\midrule
Ours& \textbf{47.52} & \textbf{0.031} & \textbf{0.983} \\
Only Synthetic& 44.52 & 0.081 & 0.967 \\
Only BurstSR& 47.14 & 0.037 & 0.981 \\\bottomrule
\end{tabular}

	}\vspace{1mm}%
	\caption{Impact of fine-tuning on real data}
	\label{tab:finetune}%
	\vspace{-2mm}
\end{table}

\section{Impact of input shifts}
\label{sec:impact_shift}
Here, we investigate the importance of having sub-pixel shifts in the input images for MFSR. We train and evaluate a baseline network \textbf{No Shifts} on synthetic bursts generated without any simulated camera motion. That is, all the images in the burst are identical except having different independent noise. We also include our SISR baseline for comparison. While the \textbf{No Shifts} network can exploit the burst images in order to obtain better denoising, its performance improvement over the SISR baseline is limited to $<0.6$ dB (see Table~\ref{tab:noshift}). In contrast, our approach obtains a significant improvement of $1.61$ dB in PSNR over \textbf{No Shifts} when operating on burst with sub-pixel shifts. These results show that the majority of performance gains of our approach over the SISR baseline is obtained by effective fusion of information contained in the different aliased samplings of the scene. 

\section{Impact of training dataset} 
\label{sec:impact_dataset}
We analyse the impact of pre-training our model on the synthetic data, as well as fine-tuning on the real data. We compare our approach with two baselines, i) a network \textbf{Only Synthetic} trained using only the synthetic data, and ii) a network \textbf{Only BurstSR} trained using only the real-world BurstSR dataset. The results on the BurstSR validation set are shown in Table~\ref{tab:finetune}. The network trained only using synthetic data fails to generalize to the real world images, obtaining a PSNR of $44.52$ dB. In contrast, the network trained from scratch on BurstSR performs much better with a PSNR of $47.14$ dB. The best results are obtained when combining both the strategies: pre-training first using large scale synthetic data, and finetuning the resulting network on real data. 

\section{Impact of our training loss}
\label{sec:loss}
\begin{figure*}[t]
	\centering%
	\includegraphics[trim = 0 0 0 0, width=0.8\textwidth]{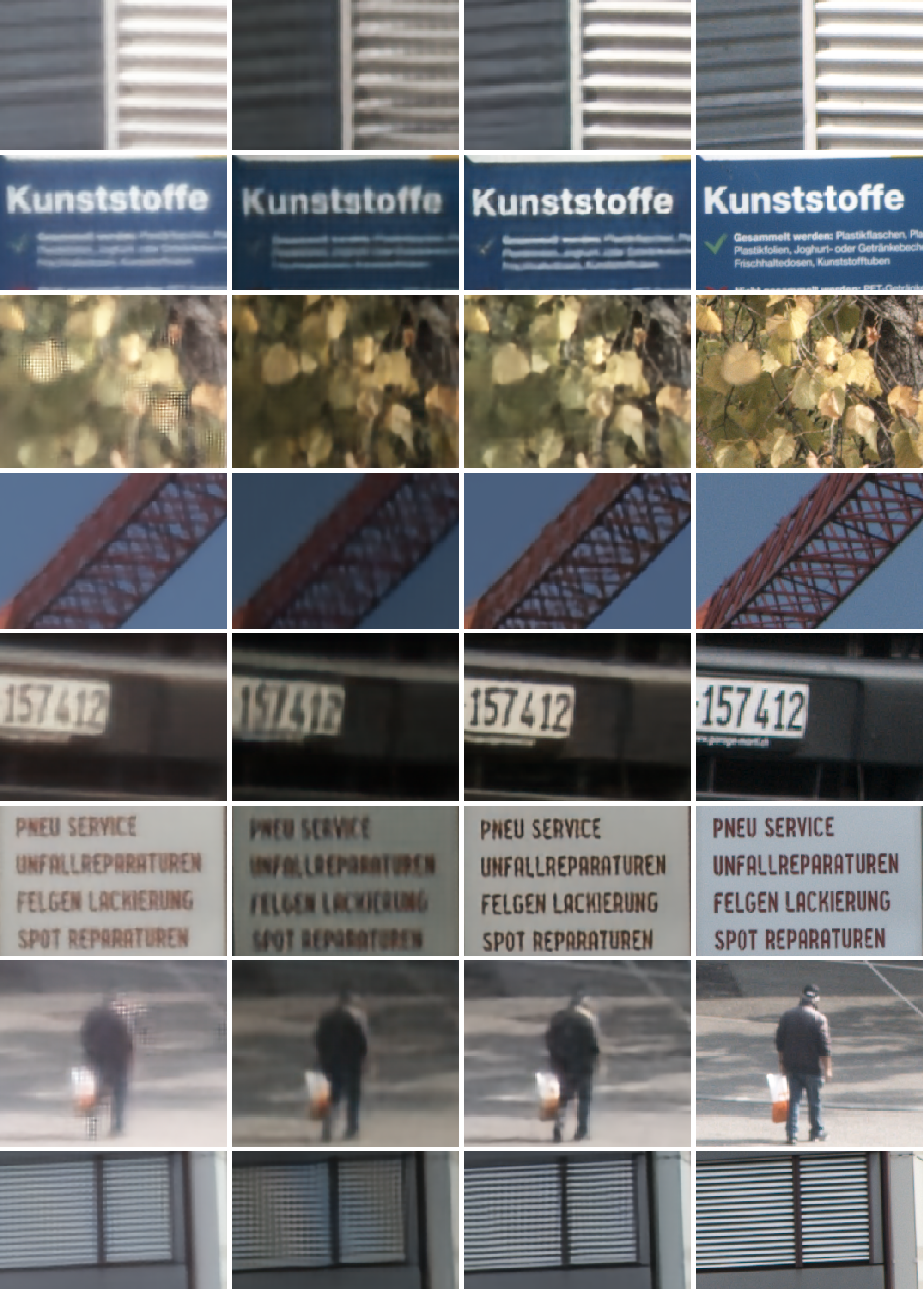}\\
	\includegraphics[trim = 0 25 0 0, width=0.8\textwidth]{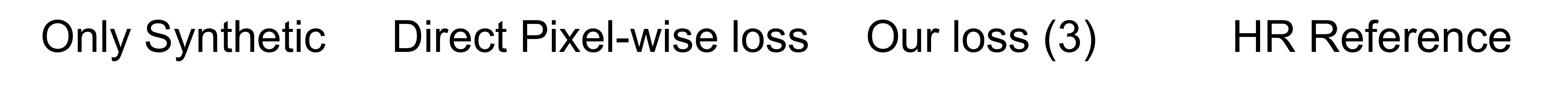}
	\caption{Qualitative comparison of a network trained on BurstSR dataset using our training loss (3) with a network trained using direct pixel-wise loss on the BurstSR validation set. A network trained only on the synthetic dataset is also included for comparison. Note that there is a color shift between the predictions of the networks, as the networks are trained using different output color spaces. Hence, we encourage the reader to focus on image details, \eg sharp edges, presence of artifacts and not on the color space differences.}\vspace{0mm}
	\label{fig:ccm_qual_fig}
\end{figure*}

In this section, we analyze the impact of our training loss, defined in Eq.~(3) in the main paper, which is used to train our model on the real-world BurstSR dataset. Our loss aligns the network prediction to the ground truth image in order to handle spatial misalignments between the input burst and the ground truth. Furthermore, it also handles the color mismatch between the input-ground truth pair by estimating the color mapping function between the two. We compare the network trained using our loss (3) with a network trained using direct pixel-wise loss without performing any explicit spatial alignment and color space correction. Additionally, we also include a network trained only on synthetic data for comparison. The results of this analysis on the BurstSR validation set are shown in Figure~\ref{fig:ccm_qual_fig}. 
Compared to using direct pixel-wise loss, the network training using our loss (3) can generate sharper images with better details. 

\section{Qualitative Examples}
\label{sec:qual}
\begin{figure*}[t]
	\centering%
	\includegraphics[trim = 0 0 0 0, width=\textwidth]{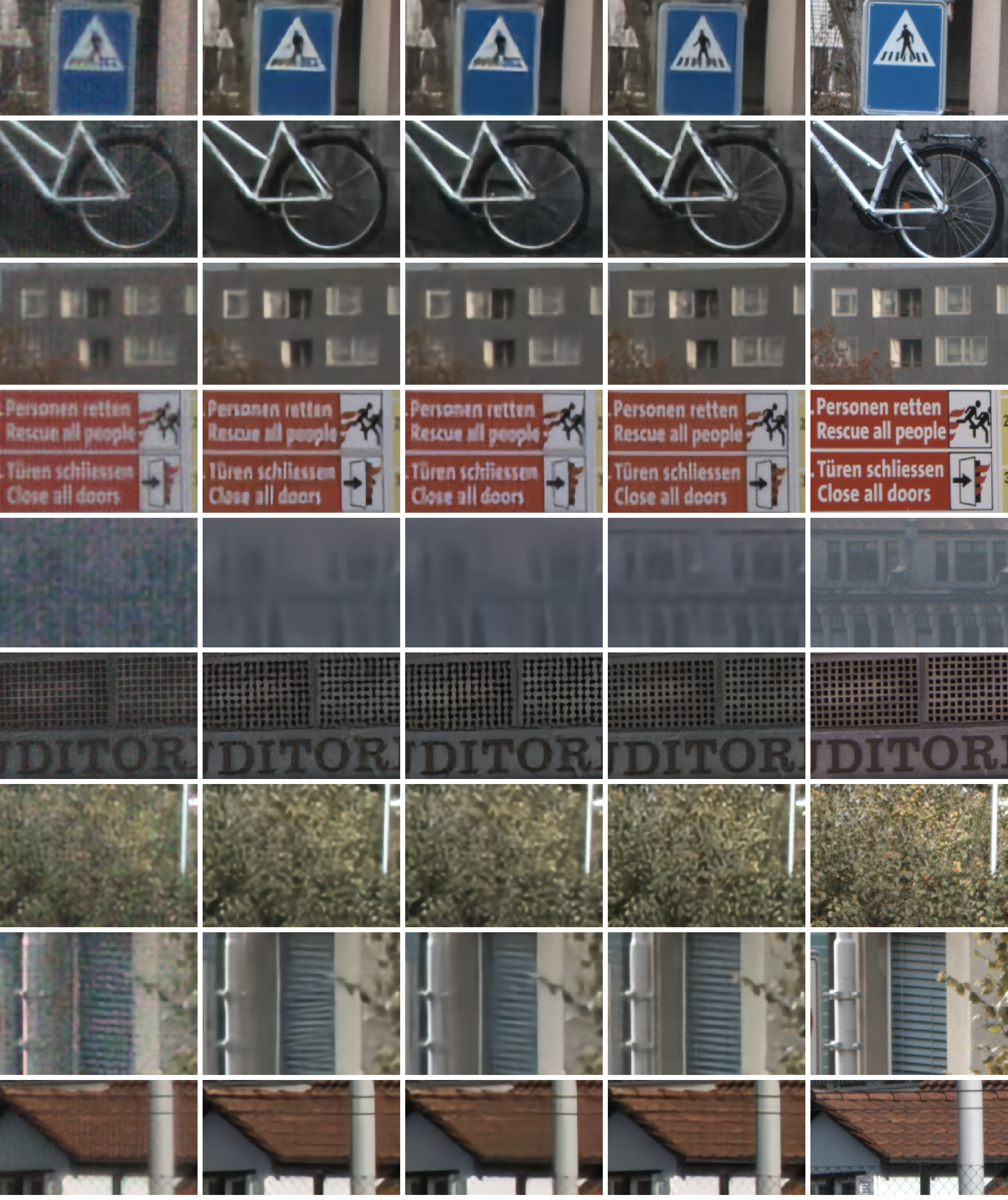}\\
	\includegraphics[trim = 0 25 0 0, width=\textwidth]{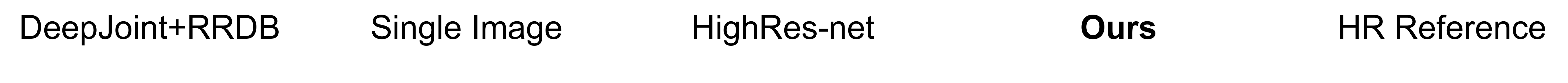}
	\caption{Qualitative comparison of our approach with existing super-resolution approaches on the BurstSR test set.}\vspace{0mm}
	\label{fig:qual_fig_supp}
\end{figure*}

Here, we provide additional qualitative comparison of our approach with the approaches described in Section 6.2 of the main paper; (i) Single Image baseline, (ii) DeepJoint~\cite{Gharbi2016DeepJD}+RRDB~\cite{Wang2018ESRGANES}, and (iii) HighRes-net~\cite{Deudon2020HighResnetRF}. Visual examples from the BurstSR test set are shown in Figure \ref{fig:qual_fig_supp}. Compared to the other methods, our approach can best reconstruct the high frequency image details with high fidelity to the high-resolution ground truth.

\end{document}